\documentclass[a4paper,fleqn]{cas-dc}

\usepackage{amssymb}
\usepackage{amsmath}
\usepackage{cleveref}
\usepackage{url}
\usepackage{xcolor}
\usepackage[normalem]{ulem}
\usepackage{booktabs}
\usepackage{pbalance}
\usepackage{longtable}
\usepackage{array}
\usepackage[toc]{appendix}
\usepackage{tabularx}
\usepackage{listings}
\usepackage{adjustbox}
\usepackage{float}

\restylefloat{table}

\crefrangelabelformat{equation}{#3#1#4--#5#2#6}
\crefname{equation}{axiom}{axioms}
\Crefname{equation}{Axiom}{Axioms}

\begin{document}



\let\WriteBookmarks\relax
\def\floatpagepagefraction{1}
\def\textpagefraction{.001}
\shorttitle{The BDI Ontology}
\shortauthors{S. Zuppiroli et al.}

\title [mode = title]{The Belief-Desire-Intention Ontology for modelling mental reality and agency}                      

\author[ISTC]{Sara Zuppiroli}
    [orcid=0009-0006-3298-6232]
    \ead{sara.zuppiroli@cnr.it}

\author[ISTC]{Carmelo Fabio Longo}
    [orcid=0000-0002-2536-8659]
    \ead{carmelofabio.longo@cnr.it}

\author[ISTC,UNIBO]{Anna Sofia Lippolis}
    [orcid=0000-0002-0266-3452
]
    \ead{annasofila.lippolis@istc.cnr.it}

\author[IRPPS]{Rocco Paolillo}
    [orcid=0000-0001-9816-5839]
    \ead{rocco.paolillo@cnr.it}

\author[IRCRES]{Lorenzo Giammei}
    [orcid=0009-0009-3976-0107]
    \ead{lorenzo.giammei@cnr.it}

\author[ISTC]{Miguel Ceriani}
    [orcid=0000-0002-5074-2112]
    \ead{miguel.ceriani@cnr.it}

\author[ISTC]{Francesco Poggi}
    [orcid=0000-0001-6577-5606]
    \ead{francesco.poggi@cnr.it}

\author[IRCRES]{Antonio Zinilli}
    [orcid=0000-0001-8505-5040]
    \ead{antonio.zinilli@cnr.it}
    
\author[ISTC]{Andrea Giovanni Nuzzolese}
    [orcid=0000-0003-2928-9496]
    \ead{andreagiovanni.nuzzolese@cnr.it}

\affiliation[ISTC]{organization={CNR - Institute of Cognitive Sciences and Technologies},
            city={Bologna, Rome \& Catania},
            country={Italy}}

\affiliation[IRPPS]{organization={CNR - Institute for Research on Population and Social Policies},
            city={Rome},
            country={Italy}}

\affiliation[IRCRES]{organization={CNR - Research Institute on Sustainable Economic Growth},
            city={Rome},
            country={Italy}}
            
\affiliation[UNIBO]{organization={University of Bologna - Department of Philosophy},
            city={Bologna},
            country={Italy}}

\begin{abstract}
The Belief-Desire-Intention (BDI) model is a cornerstone for representing rational agency in artificial intelligence and cognitive sciences. Yet, its integration into structured, semantically interoperable knowledge representations remains limited. This paper presents a formal BDI Ontology, conceived as a modular Ontology Design Pattern (ODP) that captures the cognitive architecture of agents through beliefs, desires, intentions, and their dynamic interrelations. The ontology ensures semantic precision and reusability by aligning with foundational ontologies and best practices in modular design. Two complementary lines of experimentation demonstrate its applicability: (i) coupling the ontology with Large Language Models (LLMs) via Logic Augmented Generation (LAG) to assess the contribution of ontological grounding to inferential coherence and consistency; and (ii) integrating the ontology within the \textsc{Semas} reasoning platform, which implements the Triples-to-Beliefs-to-Triples (T2B2T) paradigm, enabling a bidirectional flow between RDF triples and agent mental states. Together, these experiments illustrate how the BDI Ontology acts as both a conceptual and operational bridge between declarative and procedural intelligence, paving the way for cognitively grounded, explainable, and semantically interoperable multi-agent and neuro-symbolic systems operating within the Web of Data.
\end{abstract}

\begin{keywords}
Ontology Design Patterns \sep eXtreme Design \sep Ontology Engineering \sep Belief-Desire-Intention \sep Mental models
\end{keywords}

\maketitle

\section{Introduction}
\label{sec:intro}

The Belief-Desire-Intention (BDI) model~\cite{bratman1987intention} deserves particular attention due to its fundamental role in representing and reasoning about agency. Since the early 2000s, significant efforts have been devoted to developing open multi-agent systems across a range of application domains, often involving heterogeneous participants designed by independent teams~\cite{luck2003agent}. Numerous architectures for implementing agent-based systems have been proposed in the literature, such as those reviewed in~\cite{de2020bdi}. Likewise, BDI-based approaches have found applications in various fields, including the Web, social sciences, and economics~\cite{adam2016bdi}.

Moreover, BDI agents mark a significant milestone in realising the {\em Cognitive Era} of Artificial Intelligence (AI)~\cite{bordini2020agent}, offering an ideal framework for integrating the diverse capabilities required to advance towards the next generation of cognitively grounded intelligent systems. The growing demand for explainable, interpretable, and human-aligned AI has renewed interest in hybrid approaches that combine symbolic knowledge with statistical reasoning. Within this landscape, the BDI model has emerged as a cornerstone for representing rational agency, providing a structured account of how agents perceive, deliberate, and act in dynamic environments. However, while BDI models offer strong conceptual foundations, they often lack the semantic interoperability and scalability needed to operate effectively within the Web of Data and modern neuro-symbolic ecosystems. To address this limitation, this paper introduces a formal {\em BDI Ontology}, designed to capture the cognitive architecture of intelligent agents while ensuring compliance with Semantic Web standards and ontological best practices.

In this context, the BDI Ontology provides a shared conceptual vocabulary for representing mental states (i.e. beliefs, desires, and intentions), goals, plans, and justifications along with their dynamic relations. It enables explicit modelling of how agents form, revise, and reason over these states, thus making deliberative processes transparent and machine-interpretable. Through this formalisation, the ontology supports the integration of symbolic reasoning within heterogeneous computational settings, including large language models (LLMs) and logic-based BDI frameworks. The result is a unified framework that links the semantic representation of agency with executable reasoning, fostering explainability and reproducibility across AI systems as envisioned by~\cite{Huang2024}.

Additionally, the paper presents two main lines of experimentation. The first explores Logic Augmented Generation~\cite{Gangemi2025} (LAG) to couple the BDI ontology with LLMs to evaluate their inferential and modelling abilities when prompted with formal conceptual knowledge. Through the use of the MS-LaTTE dataset~\cite{jauhar2022ms}, the study assesses whether ontological grounding enhances the model’s capability to detect logical inconsistencies, generate coherent mental-state representations, and answer the ontology’s competency questions. 

The second line of experimentation focuses on the integration of the BDI ontology into executable BDI frameworks, specifically within \textsc{Semas}, a Prolog-style reasoning platform implementing the Triples-to-Beliefs-to-Triples (T2B2T) paradigm. In this setting, the ontology is actively used to represent and manage agents’ internal knowledge, including beliefs, desires, intentions, and their temporal evolution. It supports querying and reasoning over agents’ mental states, thereby enabling the operational use of the ontology within a running agent system rather than as a purely conceptual model. 

Together, these contributions demonstrate how the BDI ontology serves as both a conceptual and operational bridge between declarative and procedural intelligence. It empowers neuro-symbolic and agent-based systems to reason coherently about their beliefs, desires, and intentions, while maintaining semantic alignment with shared knowledge graphs. In doing so, the ontology lays the groundwork for cognitively plausible, explainable, and semantically interoperable AI agents capable of perceiving, reasoning, and acting consistently within the evolving Web of Data.

The paper is organised as follows: Section~\ref{sec:related} provides an overview of the current state of the art; Section~\ref{sec:onto} presents the ontology model; Section~\ref{sec:app} offers examples of applications of our model in practice. Finally, Section~\ref{sec:conclusions} concludes the paper and provides future work.

\section{Background}
\label{sec:related}
In the next section we first provide the background on the BDI model (cf. Section~\ref{sec:related-literature}), then we identify existing ontologies for the BDI model (cf. Section~\ref{sec:related-ontologies}).
\subsection{Belief-Desire-Intention: a model for representing agency}
\label{sec:related-literature}
The concept of {\em agency}, in a broad sense, refers to the capacity of an entity to act intentionally and autonomously in pursuit of goals.
Causal theories in philosophy have significantly informed the study of agents in artificial intelligence. An example of this approach is the belief–desire–intention (BDI) theory, formulated by the philosopher Michael Bratman in the 1980s. 
In ~\cite{bratman1987intention}, Bratman describes a theory of human practical reasoning that laid the foundations for modelling intentional agency, both in philosophy and in artificial intelligence. Within this theory, the \emph{mental states} of \emph{Belief}, \emph{Desire}, and \emph{Intention} (BDI) correspond, respectively, to the informational, motivational, and deliberative dimensions of agency. 
Bratman argues that it is intention that causes action. Thus, the causal sequence leading an agent to act begins with the understanding of desires and the knowledge of beliefs that enable the articulation of intentions, but it is the intentions that directly cause the action. In the field of automated agents, Rao et al. \cite{rao1995bdi} developed a more practical system for rational reasoning. Their model builds on the Procedural Reasoning System (PRS) \cite{georgeff, ingrand}, one of the earliest implemented agent-oriented systems based on the BDI architecture. The BDI architecture was later extended in the successor system dMARS \cite{DInverno2004} (Distributed Multi-Agent Reasoning System). They represent only beliefs about \emph{current} state of the world (which can be expected to change over time), by considering only ground sets of literals with no disjunctions or implications, where the so-defined \emph{plans} are considered a special form of beliefs used as means of achieving certain future \emph{world} states. 
\par In general, BDI logic relates actions to subsets of possible worlds, which represent the beliefs of an agent. Such belief states can be modelled using Decision Trees (DT). A DT is composed of decision nodes, chance nodes, and terminal nodes, and is associated with both a probability function mapping chance nodes to probabilities, and a payoff function mapping terminal nodes to real values. A deliberation function then selects one or more optimal sequences of actions for a given node.
The practical deployment of multi-agent systems (MAS) in industrial environments has introduced new challenges, particularly in achieving interoperability across heterogeneous platforms. As automation solutions from different vendors began to coexist, the absence of shared communication standards quickly became a critical bottleneck—an issue further amplified by the rise of the Internet of Things (IoT). Early industrial initiatives often favoured proprietary agent-based systems, where devices produced by a single manufacturer could communicate only within their own technological ecosystem. This fragmentation underscored the need for a horizontal, standardised approach to agent communication, as opposed to isolated, vendor-specific architectures, to ensure seamless interaction among diverse agents and services.

In response to this need, the Foundation for Intelligent Physical Agents (FIPA)\footnote{\url{http://www.fipa.org/}}
 was established as an international consortium to develop specifications fostering interoperability among heterogeneous agents. Several widely adopted MAS frameworks—such as JACK\footnote{\url{https://aosgrp.com.au/jack/}}
, JADE\footnote{\url{https://jade.tilab.com/}}
, JADEX \cite{10.1007/3-7643-7348-2_7}, and JaCaMo \cite{BOISSIER2013747}—adhere to FIPA standards, offering reference implementations that demonstrate the practical value of such interoperability principles.

\subsection{Ontologies for modelling BDI}
\label{sec:related-ontologies}
The adoption of ontologies in Belief–Desire–Intention (BDI) agents~\cite{rao1995bdi} and, more generally, across Multi-Agent Systems (MAS) has become a well-established and widely adopted practice. These applications can be divided into three groups:
(i) Ontologies as sources for knowledge base derivation; (ii) Ontologies supporting model-driven engineering; and (iii) Ontologies in BDI-based agent programming languages.

In the first group, we find works such as ~\cite{holmes2009augmenting} and~\cite{Busetta1999}, where the authors use a set of ontologies to derive agent-specific knowledge bases. For example, in the case of JACK agents~\cite{Busetta1999}, which is a framework in Java for multi-agent system development, concepts (i.e. {\em belief sets}) are read directly from specified concepts and individuals in the ontology. Similarly, in \cite{dickinson2005agents}, the authors employ the Nuin agent platform \cite{dickinson2003towards}, an open-source Java application that combines belief-desire-intention agents with semantic web techniques.

In the second group, ontologies are employed to support model-driven engineering approaches, where the design of multi-agent systems is derived by instantiating domain ontologies, thus ensuring semantic alignment between specification and implementation \cite{freitas2017model}.

The third group comprises works that aim to extend the knowledge representation capabilities of Agent Programming Languages used in BDI-based systems. One notable direction focuses on the integration of ontological formalisms. For example, Moreira et al.~\cite{moreira2006agent} propose an extension of the BDI agent-oriented programming language AgentSpeak, grounded in Description Logic (DL). In their approach, the agent's belief base encompasses both definitions of complex conceptual structures and specific factual knowledge, thereby enhancing the expressiveness and reasoning capabilities of BDI agents. The advantages of integrating AgentSpeak with DL include: (i) queries to the belief base are more expressive, as results do not rely solely on explicit knowledge but can be inferred from the ontology;
(ii) the notion of belief update is refined since the (ontological) consistency of a belief addition can be verified;
(iii) the search for a plan to handle an event becomes more flexible, as it is not based solely on unification but on the subsumption relation between concepts;
(iv) agents can share knowledge by using ontology languages such as OWL (Ontology Web Language).

Instead the authors in~\cite{klapiscak2008jasdl} introduce JASDL, an extension of the Jason agent platform that integrates ontological reasoning through the use of the OWL API. This extension enables advanced features such as plan-trigger generalisation based on ontological hierarchies and the use of ontology-based knowledge when querying the agent’s belief base.

A complementary direction aims to decouple the agent programming language from any specific knowledge representation formalism, making the underlying reasoning technology a pluggable component. This approach is exemplified by~\cite{bagosi2015designing} and~\cite{dastani2009combining}, which propose flexible architectures allowing different KR systems to be integrated seamlessly into BDI agents.

Further relevant contributions include~\cite{ferrario2005towards}, which presents a preliminary characterisation of mental categories derived from the foundational ontology DOLCE (Descriptive Ontology for Linguistic and Cognitive Engineering)~\cite{masolo2002wonderweb}, leading to the definition of the Computational Ontology of Mind (COM). Additionally, Toyoshima et al.~\cite{toyoshima2020foundations} provide both conceptual and formal foundations for an ontology of BDI, contributing to the systematic ontological grounding of cognitive architectures.

A preliminary version of the BDI ontology presented in this work was presented in~\cite{Longo2025}. both that contribution and the BDI ontology described here should be regarded as subsequent developments building upon that earlier work.

While existing approaches primarily focus on embedding ontological reasoning within specific BDI programming frameworks or proposing conceptual foundations for mental-state modelling, they generally remain tied to particular implementation environments or lack a unified formalisation of BDI constructs. In contrast, our BDI Ontology provides a domain-independent, interoperable, and semantically grounded representation of beliefs, desires, intentions, and their dynamic interrelations. Beyond its role as a reference model, the ontology is conceived as an ontology design pattern (ODP) that can be reused to support modular ontology engineering in the context of agent-based systems. This pattern-based design enables developers to extend or specialise the ontology to represent domain-specific forms of agency while maintaining conceptual consistency and interoperability across heterogeneous systems, from symbolic agent platforms to neuro-symbolic and Semantic Web environments. A preliminary version of the BDI ontology was introduced in~\cite{Longo2025} in the context of designing a system capable of translating RDF triples into agent-specific beliefs within a production rule framework. The present work substantially extends and refines that initial proposal, providing a more comprehensive formalisation and broadening its scope and applicability.

\section{The BDI ontology}
\label{sec:onto}
The BDI model, as outlined in the preceding section, has served as a foundational basis for a considerable body of research on architectures for autonomous agents. The conceptual underpinnings of this model can be traced back to the philosophical contributions of Bratman~\cite{bratman1987intention}, whose work served as a seminal source of inspiration for its development.

\subsection{Ontology requirements}

\begin{table*}[!htb]
\centering
\caption{Competency Questions for the BDI ODP}
\label{tab:cqs-bdi}
\resizebox{.8\textwidth}{!}{
\begin{tabular}{l p{11cm}}
\toprule
\textbf{ID} & \textbf{Competency Question} \\
\midrule
\multicolumn{2}{c}{\em 1. World, agents and mental entities} \\
\midrule
CQ1  & What are mental entities? \\
CQ2  & What mental states (i.e. befiefs, desires, and intentions) does an agent hold? \\
CQ3 & What are the constituent mental entities that form part of a given mental entity? \\
CQ4  & What mental processes has an agent undergone? \\
CQ5  & What is the world state that a given mental state is about? \\
\midrule
\multicolumn{2}{c}{\em 2. Dynamics of mental states} \\
\midrule
CQ6  & What beliefs motivated the formation of a given desire? \\
CQ7  & Which desire does a particular intention fulfil? \\
CQ8  & Which mental process generated a given belief, desire, or intention? \\
CQ9 & When was a mental entity generated? \\
CQ10  & What triggered a mental process? \\
CQ11 & What justifications support a specific mental entity? \\
\midrule
\multicolumn{2}{c}{\em 3. Goals and planning} \\
\midrule
CQ12 & What goal does a given intention or plan aim to fulfil? \\
CQ13 & What plan has been specified by a particular intention? \\
CQ14 & What planning process led to the creation of a particular plan? \\
CQ15 & What is the ordered sequence of tasks that compose a given plan? \\
\midrule
\multicolumn{2}{c}{\em 4.    Temporal reasoning} \\
\midrule
CQ16 & What is the temporal validity (start and end time) of a mental state? \\
CQ17 & What mental states were valid at a specific point in time?  \\
CQ18 & How has a mental entity evolved over time? \\
\bottomrule
\end{tabular}
}
\end{table*}

In this work, we  model the mental reality of agents in terms of mental states and mental processes. To address these objectives, we identified a set of functional requirements for the ontology design process. These requirements are formalised as {\em competency questions} (CQs)~\cite{Gruninger1995}, a widely adopted approach to represent requirements in various ontology engineering methodologies, including {\em eXtreme Design} (XD)~\cite{Blomqvist2010}, which we adopted in this work. 
The eXtreme Design (XD) methodology is an agile, iterative approach to ontology engineering that emphasises the reuse and composition of Ontology Design Patterns~\cite{Gangemi2009} (ODPs). Inspired by agile software development practices, XD structures ontology development around short, collaborative design cycles in which competency questions guide the identification, selection, and adaptation of suitable patterns from existing repositories. Each iteration involves analysing requirements, choosing relevant ODPs, integrating and testing them, and refining the ontology based on feedback. This process promotes modularity, reusability, and rapid prototyping, allowing ontology engineers to build complex models incrementally while ensuring semantic coherence and alignment with established design best practices.
Following the XD methodology, we analysed the literature on BDI, as discussed in Section~\ref{sec:related}, and engaged with experts in the domain of agent-based modelling to formulate the competency questions listed in Table~\ref{tab:cqs-bdi}.
Following the XD methodology, we first analysed the literature on BDI, as discussed in Section~\ref{sec:related}, to ensure alignment with established theoretical foundations. Building on this, and in close collaboration with domain experts, we developed a set of user stories. These user stories, reported in Appendix A, formed the basis for deriving the competency questions presented in Table~\ref{tab:cqs-bdi}.
The competency questions were formulated following a structured approach aligned with the XD methodology. Starting from the elicited user stories, we identified the underlying informational needs by analysing the actors, goals, and contextual constraints expressed in each scenario. These needs were then reformulated as natural language questions, capturing the key queries that the ontology should be able to answer. Each candidate question was subsequently mapped to the relevant domain concepts and relations, ensuring consistency with the conceptual model and the terminology emerging from the literature. The questions were iteratively refined to guarantee clarity, atomicity, and non-redundancy, and were validated in collaboration with domain experts to ensure their relevance and completeness. This process ensured that the final set of competency questions provides a faithful operationalisation of the requirements captured by the user stories, while remaining aligned with the theoretical foundations of the BDI paradigm.
The CQs are thematically organised to reflect key dimensions of the BDI architecture. The first group of CQs, {\em World, agents and mental entities}, concerns the association between agents and their internal cognitive states and processes, which are referred to as {\em mental entities}. The underlying modelling rationale here is to treat agents as entities capable of holding or participating in mental states such as beliefs, desires, and intentions. These states are considered cognitive attributes of agents and are typically reified as entities that persist over time, i.e. endurants. The mental processes involved in their formation (e.g. belief formation) are treated as temporally extended events, i.e. perdurants, enabling explicit reasoning about agent dynamics and transitions between states. These distinctions are fairly represented in DOLCE UltraLight that provides concepts for modelling agents, events, and situations. 
Consequently, these design patterns have been selected for reuse: (i) the \texttt{EventCore} pattern\footnote{\url{https://odpa.github.io/patterns-repository/EventCore/EventCore}}, which supports the modelling of events that occur in time and involve participants; (ii) the \texttt{Situation} pattern\footnote{\url{https://odpa.github.io/patterns-repository/Situation/Situation}} provides a way to encapsulate a configuration of entities holding at a given time. 
For example, our BDI ontology aims at modelling events to represent mental processes such as intention formation, where an agent commits to a goal after deliberation. Similarly, a belief can be represented as a situation comprising a proposition and a perceived world state at a particular time.

Additionally, the first group of CQs addresses the intentional stance of mental entities, capturing their semantic alignment with world states and their underlying justifications. In this context, a world state can be understood as analogous to the notion of a layout introduced in~\cite{masolo2002wonderweb}, that is, a structured configuration of entities and conditions that characterise a particular situation independently of any agent’s perspective. However, in the present ontology, the world state is not just an abstract possibility, but a contextualised situation that describes the environment regardless of agents. It serves as the referential substrate upon which agents construct their beliefs, formulate desires, and adopt intentions. Mental states are thus directed toward these world states, which provide the semantic ground for intentionality. Furthermore, the relationship between a mental state and its supporting justifications is modelled to account for the epistemic or motivational basis on which agents adopt such states, e.g. evidence supporting a belief, or a goal motivating a desire. The \texttt{Provenance}\footnote{\url{https://odpa.github.io/patterns-repository/Provenance/Provenance}} ODP has been selected as a reusable modelling solution to address the latter scenario.

The {\em Dynamics of mental states} group of CQs addresses the causal and inferential mechanisms underlying the emergence and transformation of mental entities within the BDI architecture. These questions aim to capture how beliefs, desires, and intentions are not static or isolated constructs, but are dynamically interrelated through processes of reasoning, motivation, and commitment. From an ontological standpoint, mental states are treated as endurants, while the cognitive activities that produce or modify them (e.g., belief formation, intention adoption) are modelled as perdurants. This modelling approach supports the explicit representation of causal dependencies. For example, a belief may motivate the formation of a desire, which in turn gives rise to an intention. Such transitions are grounded in an ontological understanding of agentive causality, where one mental entity serves as a contributing factor in the generation of another, mediated through a mental process. These relationships are not merely correlational, but encode a form of internal deliberative logic—tracing the agent's reasoning path from perception to motivation to commitment. Again, the \texttt{Provenance} ODP serves as the foundational pattern for modelling these aspects in our ontology.

The {\em Goals and Planning} group of CQs articulates the operational dimension of the BDI model, specifically the ontological commitments required to represent the link between intentions, plans, and goals. In this framework, goals are modelled as desirable future states (i.e. configurations of the world that agents aim to bring about), whilst intentions reflect the agent's commitment to achieving those goals. This commitment entails not only a motivational stance but also the adoption of a concrete means for realisation, typically encoded as a plan. Plans are understood as structured sequences of actions or procedures, often constrained by context and feasibility. Ontologically, this reflects {\em telicity}, a notion rooted in Aristotelian philosophy, which concerns whether an event is characterised by an inherent or intended endpoint, i.e. its telos. Our representation of goals and planning adopts this perspective, aligning with the direction recently outlined by~\cite{Guarino2024}, where intentions and plans are characterised by an agent's commitment to achieving a specific telos that guides and constrains deliberative behaviour.
Additionally, this aligns with classical planning theory and finds ontological grounding in patterns that capture goal realisation, action specification, and fulfilment conditions, such as the \texttt{BasicPlan} ODP\footnote{\url{https://odpa.github.io/patterns-repository/BasicPlan/BasicPlan}}, which we have adopted for reuse. Such an ODP enables us to model whether a plan fulfils a given goal, or which goal an intention operationalises, thereby supporting fine-grained reasoning over deliberative behaviour and strategic decision-making.

The {\em Temporal Reasoning} group supports the explicit representation of temporal dimensions associated with mental states and processes. Mental states such as beliefs, desires, and intentions are not static; instead, they emerge, persist, and evolve over time. Similarly, the events that generate these states (e.g. deliberation, perception, or communication) are inherently temporal in nature. This requires the ontology to distinguish between instantaneous and durative entities, and to associate both with temporal intervals or instants. Drawing on foundational ontologies such as DOLCE \cite{Gangemi2002}, the model represents temporal validity via time-indexed situations and treats cognitive activities as perdurants that unfold over time. This allows for diachronic reasoning such as querying which mental states were active at a given moment, or how an agent’s intention evolved in response to new beliefs. Temporal reasoning is thus essential for tracking the lifecycle of mental entities, coordinating actions, and evaluating plan execution within dynamically changing environments. The \texttt{TimeIndexedSituation} ODP\footnote{\url{https://odpa.github.io/patterns-repository/TimeIndexedSituation/TimeIndexedSituation}} has been selected to model the requirements associated with temporal reasoning.

Together, these groups of questions define not only the representational scope of the ontology, but also its inferential utility. They ensure that the model supports reasoning across epistemic (belief-based), motivational (desire- and goal-based), and temporal dimensions. This enables the agents to be described not only in terms of what they think and want, but when and why those states emerge and evolve. This pattern-based representational approach is essential for modelling deliberative agency in a way that is both semantically precise and computationally actionable.

\subsection{Ontology formalisation (overview)}

\begin{figure*}
\centering
\includegraphics[scale=0.38]{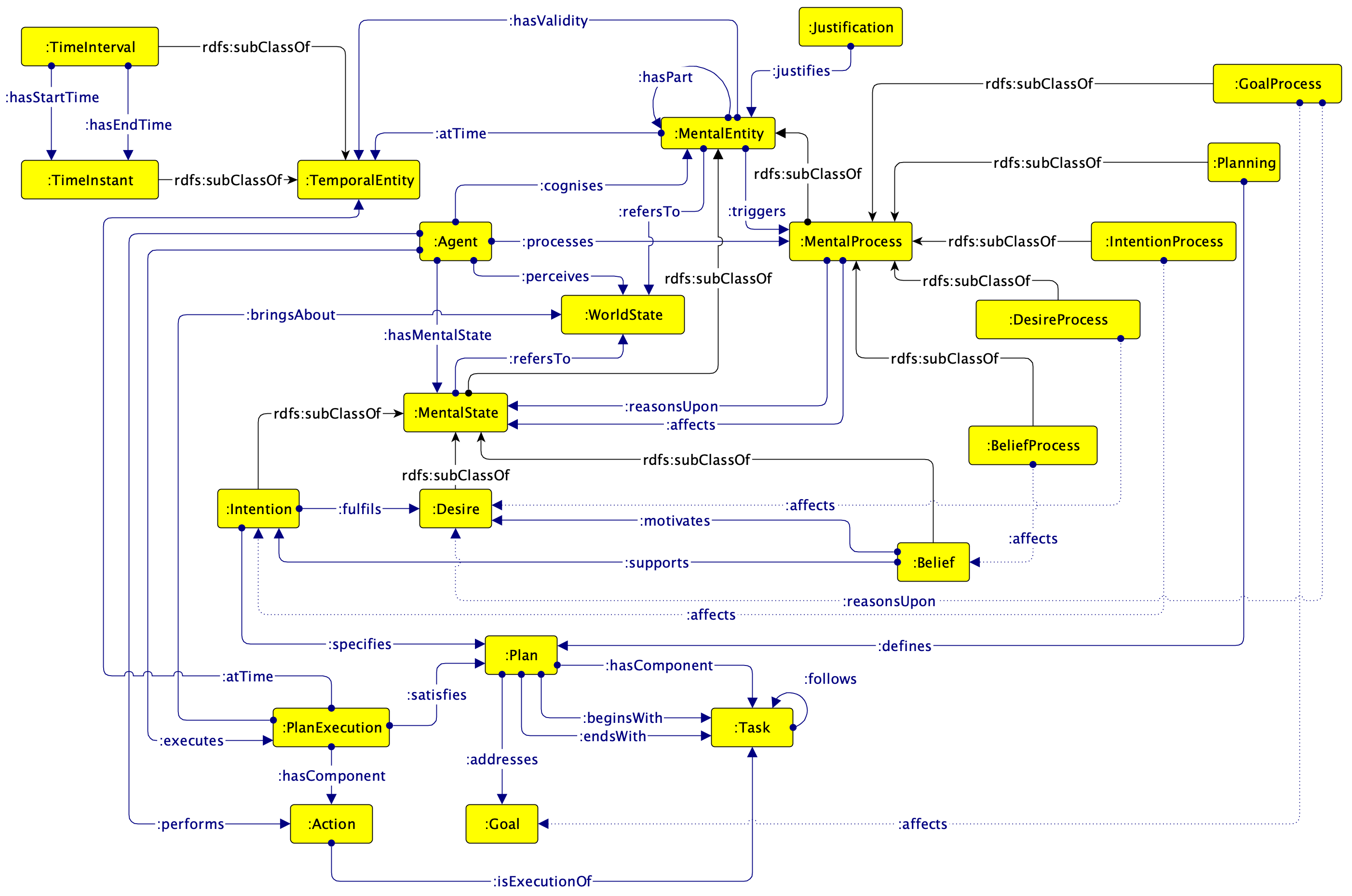}
\caption{The \textsc{bdi} ontology.}
\label{fig:schema_bdi}
\end{figure*}

Our formalisation of the BDI ongology provides a formal framework for modelling the Belief-Desire-Intention (BDI) architecture for rational agents. For this, it defines key mental states and their relationships, thereby capturing the reasoning, motivation, and commitment to action of the agent. Figure~\ref{fig:schema_bdi} shows the Graffoo~\cite{falco2014modelling} diagram of the BDI ontology.  
The ontology is grounded in DOLCE-0\footnote{DOLCE-0 is a supplementary ontology used as a generalisation of DOLCE+DnS Ultralite. DOLCE-0 is available online at \url{http://www.ontologydesignpatterns.org/ont/d0.owl}.}.


The detailed formalisation of the ontology, including the full set of axioms, is reported in Appendix~\ref{sec:app2}.

\paragraph{\bf World, agents and mental entities.}

The foundational layer of the BDI ontology establishes the core 
classes and relations needed to represent agents, their cognitive 
content, and the world they inhabit. The class \texttt{:WorldState}\footnote{The prefix \texttt{:} stands for the namespace \url{https://w3id.org/fossr/ontology/bdi/}} 
captures a temporally situated condition of the environment as 
perceived by an agent, grounding the agent's reasoning in observable 
external facts. It is aligned with \texttt{d0:Eventuality} from 
DOLCE-0, and linked to agents via the inverse property pair 
\texttt{:perceives}/\texttt{:isPerceivedBy}, where an agent 
perceives a world state and, conversely, a world state is perceived 
by an agent. The class \texttt{:Agent}, a subclass of 
\texttt{dul:Agent}, represents any autonomous entity capable of 
perceiving its environment, reasoning about it, and acting to 
achieve goals. Agents cognise mental content through the property 
\texttt{:cognises}, whose range is constrained to 
\texttt{:MentalEntity}, which is a subclass of \texttt{d0:CognitiveEntity} 
representing all cognitive constructs an agent can mentally hold. 
Mental entities are further structured via the \texttt{:hasPart} object property, which 
formalises meronymic relations among cognitive constructs, enabling 
compositional representations such as the decomposition of a belief 
into its temporal and locational components. Mental states, i.e.  
beliefs, desires, and intentions, are introduced as subclasses of 
\texttt{:MentalEntity}, each further closed under \texttt{:hasPart}. 
Mutual disjointness among \texttt{:Belief}, \texttt{:Desire}, and 
\texttt{:Intention} is enforced as a global axiom, ensuring 
ontological consistency across the cognitive layer.

\paragraph{\bf Dynamics of mental states.}
Mental states are not static constructs: they emerge, evolve, and  are suppressed through cognitive activity. The class  \texttt{:MentalProcess}, defined as a subclass of both  \texttt{:MentalEntity} and \texttt{d0:Activity}, represents the  reasoning activities through which an agent updates its internal  representation of the world. Each mental process is grounded in one or more existing mental states via \texttt{:reasonsUpon}, may be triggered by any mental entity via \texttt{:isTriggeredBy}, and affects mental states through the property \texttt{:affects}, which specialises into \texttt{:generates}, \texttt{:modifies}, and \texttt{:suppresses} to cover the full lifecycle of a mental state. Three specialised process types are defined, namely: (i) \texttt{:BeliefProcess}, (ii)\texttt{:DesireProcess}, and (iii) \texttt{:IntentionProcess}. Each of them is constrained to affect only the corresponding mental state type. Direct inter-state relations, i.e. \texttt{:motivates}, \texttt{:supports}, and \texttt{:fulfils}, enable the construction of structured cognitive chains, linking beliefs to desires and intentions in a traceable deliberative sequence. Explainability is further supported by the class \texttt{:Justification}, a subclass of \texttt{dul:Description}, which anchors any mental entity to the evidence or rationale that gave rise to it via \texttt{:justifies}.

\paragraph{\bf Goals and planning.}
The planning layer formalises how an agent's intentions are 
translated into actionable structures aimed at achieving explicit 
goals. The class \texttt{:Plan}, aligned with \texttt{dul:Plan}, 
represents a structured sequence of tasks designed to fulfil a 
\texttt{:Goal}, a subclass of \texttt{dul:Goal}, via 
the property \texttt{:addresses}. Intentions initiate planning 
through the property \texttt{:specifies}, establishing a direct 
link between deliberative commitment and concrete strategy. The 
class \texttt{:Planning} is modelled as a \texttt{:MentalProcess} 
that reasons exclusively upon intentions and produces plans via 
\texttt{:defines}. The internal structure of plans is captured 
through \texttt{:Task} instances, ordered by the transitive 
properties \texttt{:follows} and \texttt{:precedes}, and bounded 
by \texttt{:beginsWith} and \texttt{:endsWith}. Plan execution is 
represented by \texttt{:PlanExecution}, a subclass of 
\texttt{dul:PlanExecution}, which links the enacted plan to the 
performing agent, the resulting \texttt{:WorldState}, and a 
\texttt{:TemporalEntity}, supporting causal and temporal reasoning 
over agent behaviour. The class \texttt{:Action} captures the 
concrete realisation of individual tasks, enabling the comparison 
between intended and actual behaviour.

\paragraph{\bf Temporal reasoning.}
Every mental entity in the ontology is temporally contextualised  through two complementary solutions. In fact, on one hand the property \texttt{:atTime}  anchors a mental entity to the moment of its occurrence or generation, whilst, on the other, the property \texttt{:hasValidity} captures the temporal interval over which it remains active or applicable. 
Temporal constructs are represented by the class \texttt{:Temporal\-Entity}, a subclass of dul:Region, further specialised into \texttt{:TimeInstant} and \texttt{:TimeInterval}. These represent, respectively, a zero-duration temporal point associated with an \texttt{xsd:dateTime} value, and a temporal interval constrained to have exactly one start time and at most one end time, thereby supporting both open and closed intervals.

\subsection{Ontology implementation and validation}
The ontology is implemented with OWL 2 and is under version control into a dedicated GitHub repository \footnote {    \url{https://w3id.org/fossr/ontology/bdi}  } .
The DL expressivity of the implemented ontology is $\mathcal{SRIQ}(\mathcal{D})$, which supports role hierarchies, inverse properties, qualified number restrictions, and datatypes. 

Table~\ref{tab:ontometrics} reports key structural and logical metrics that provide a quantitative overview of the BDI ontology. The ontology comprises a total of 547 axioms, of which 288 are logical axioms. It defines 22 classes and 71 object properties, reflecting a modelling focus on rich relational structures. Among these, 33 inverse object property axioms and 28 sub-property axioms are specified, enabling expressive role hierarchies and bidirectional navigation of relationships. The ontology includes 72 subclass axioms, 5 disjointness declarations, and supports role characteristics such as transitivity (4 axioms), asymmetry (2), and reflexivity (2), all of which contribute to enhancing reasoning precision.
Domain and range constraints are systematically specified for object properties, with 71 domain and 69 range axioms, ensuring semantic clarity and consistency in relation use. Annotation support is robust, with 167 annotation assertions that enhance human readability and documentation. Regarding quality indicators, the ontology exhibits a relatively high axiom-to-class ratio (24.87) and inheritance richness (3.27), suggesting a strong hierarchical structure.
A particularly significant indicator is the relationship richness, measured at 0.51. This metric reflects the proportion of non-inheritance relationships (i.e., object properties other than \texttt{rdfs:subClassOf}) relative to the total number of relationships in the ontology. A relationship richness greater than 0.5 suggests that the ontology encodes a diverse and semantically expressive set of inter-class relations, beyond simple taxonomic hierarchies. This is consistent with the BDI ontology’s focus on modelling cognitive dynamics, intentional behaviour, and complex agent interactions, which require more than just class subsumption to represent meaning. As such, the ontology supports detailed reasoning over agent mental states, planning processes, and world state transitions
The inverse relations ratio (0.46) and class-to-relation ratio (0.15) confirm the ontology's orientation toward relational expressivity over class proliferation, in line with its cognitive and process-oriented modelling goals.

The namespace \url{https://w3id.org/fossr/ontology/bdi} identifies the ontology and provides stable and persistent URIs for all the concepts and properties that it defines. This ensures long-term accessibility and interoperability in accordance with best practices in linked data publishing. A content negotiation mechanism has been set up so that clients accessing the ontology can retrieve it either as human-readable HTML or in one of the standard machine-readable serialisations supported by OWL (i.e., RDF/XML, Turtle, and N-Triples), depending on the request headers.

\begin{table}[!ht]
\centering
\caption{Ontology metrics.}
\label{tab:ontometrics}
\resizebox{.38\textwidth}{!}{
\begin{tabular}{lccc}
\toprule

\textbf{Metric} & \textbf{Value}       \\
\midrule
Axioms & 547 \\
Logical axioms & 288 \\
Classes & 22 \\
Object properties & 71 \\
Inverse object properties axioms & 33 \\
SubClassOf axioms & 72 \\
Disjoint classes axioms & 5 \\
SubObjectPropertyOf axioms & 28 \\
Transitive object property axioms & 4 \\
Asymmetric object property axioms & 2 \\
Reflexive object property axioms & 2 \\
Object property domain axioms & 71 \\
Object property range axioms & 69 \\
Annotation assertion axioms count & 167 \\
Attribute richness & 0.05 \\
Inheritance richness & 3.27 \\
Relationship richness & 0.51 \\
Axiom/class ratio & 24.87 \\
Inverse relations ratio & 0.46 \\
Class/relation ratio & 0.15 \\
\bottomrule
\end{tabular}
}
\end{table}

The ontology aligns with the DOLCE+DnS UltraLite (DUL) foundational ontology, leveraging its upper-level constructs such as \texttt{dul:Agent}, \texttt{dul:Plan}, \texttt{dul:Action}, and \texttt{dul:Description}. These alignments are maintained in a separate OWL file at \url{https://w3id.org/fossr/ontology/bdi-dul}, which imports both the BDI ontology and DUL. This modular approach allows users to adopt the BDI ontology either independently or in conjunction with DUL, depending on their modelling requirements. The separation of alignments follows the direct re-use strategy (Carriero et al., 2020), which recommends reusing external ontologies by importing their terms as templates or through formal alignments, while preserving the conceptual autonomy and modularity of the new ontology. It is worth clarifying that the OWL file that provides the alignments  between the BDI ontology and DUL is also under version control on the same GitHub repositoryThe\footnote{The ontology file is available at \url{https://github.com/fossr-project/ontology/blob/main/bdi-dul/bdi-dul.rdf}.}. Additionally, the ontology, its alignments, and related resources are available on Zenodo\footnote{\url{http://doi.org/10.5281/zenodo.17466326}} ensuring reproducibility and long-term availability. The resource is distributed under the Creative Commons Attribution 4.0 International (CC-BY-4.0) licence, and this information is explicitly included in the ontology using the \texttt{dcterms:license} property. Finally, the BDI ontology is currently undergoing submission for inclusion in the Linked Open Vocabularies (LOV) catalogue to further support discoverability, reuse, and community adoption.

The ontology has been annotated using OPLaX~\cite{Asprino2021} (Ontology Pattern Language eXtended), which enables the explicit association of ontology modules with their corresponding competency questions (CQs) and the related SPARQL queries used to validate them. This annotation enhances the traceability and transparency of the ontology design process by linking each conceptual element to its intended functional requirement. Furthermore, the functional commitment of the ontology has been rigorously validated using OWLUnit\footnote{\url{https://github.com/luigi-asprino/owl-unit}}, a framework for ontology unit testing that verifies the alignment between expected and actual reasoning outcomes. All unit tests, together with the corresponding SPARQL queries and validation artefacts, are openly available in the project’s GitHub repository, ensuring reproducibility and supporting community-driven evaluation and reuse.
To ensure full reproducibility and accessibility, all validation artefacts — including the complete set of SPARQL queries for all CQs, OWLUnit test cases, and example datasets — are publicly available in the project’s GitHub repository \footnote{\url{https://github.com/fossr-project/ontology/blob/main/bdi/unittesting.md}}. The repository is organised to explicitly map each CQ to its corresponding query and test, thereby providing a transparent and inspectable validation pipeline that supports reuse and independent verification. 

\section{Applications}
\label{sec:app}
We present two usage scenarios to illustrate the potential utility of the BDI ODP to support the development of cognitively grounded, explainable AI systems. The first (cf. Section~\ref{sec:lag}) explores the integration of the ontology with large language models through Logic-Augmented Generation~\cite{Gangemi2025} (LAG), showing how symbolic structures enhance the interpretability, explainability, and logical consistency of neuro-symbolic agents. The second (cf. Section~\ref{sec:semas}) discusses the ontology's integration into operational BDI frameworks, focusing on the \textsc{Semas}~\cite{Longo2025} system and the recently introduced Triples-to-Beliefs-to-Triples (T2B2T) paradigm~\cite{Longo2025}. This dual perspective shows how the BDI ontology can underpin both symbolic alignment in large-scale generative models and executable reasoning within multi-agent systems, thereby advancing a unified framework for cognitively plausible, semantically aware AI.

\subsection{Logic augmented generation application of the BDI ontology}
\label{sec:lag}
In this section, we investigate whether and to what extent the BDI ontology can provide the symbolic backbone necessary to augment a Large Language Model to a neuro-symbolic agent by improving explainability, interpretability, and cognitive plausibility of its agency. This result was achieved by evaluating two capabilities of a large language model (LLM) when prompted within a LAG setting with the BDI ontology. The first is inference, defined as the ability to detect logical inconsistencies and contradictions. The second is modelling, which refers to the ability to produce coherent and logically structured knowledge. The evaluation follows the LAG approach proposed by~\cite{Gangemi2025}. LAG is a neuro-symbolic framework that combines semantic knowledge graphs with LLMs in order to enforce logical consistency and improve the interpretability and reliability of generated outputs.
In our setting, the BDI ontology plays this symbolic role by constraining the representation of mental states and guiding the generation of RDF triples that remain consistent with a formal conceptual model. This approach is particularly valuable in domains that require explainability and factual correctness, such as agent-based systems where mental states must align with a predefined logic.

To this extent, we reuse the Microsoft Locations and Times of Task Execution (MS-LaTTE) dataset, developed by Jauhar et al. \cite{jauhar2022ms}. MS-LaTTE collects more than 10,000 to-do tasks along with the likely locations and times of day at which they are completed. In MS-LaTTE, each to-do task was annotated by 3 annotators for location and 5 annotators for time to represent the propensities of different people, and resulting diversity, in when and where tasks are most often completed. For setting up our experiment we filtered the dataset to include only annotations with high inter-annotator agreement, meaning judges must have a consensus about the inclusion of a specific location and time for a given task, applying a threshold of 0.70 for Location annotations and 0.65 for Time annotations. The lower threshold for Time reflects the greater subjectivity noted in the original paper. For this pool, we obtained a dataset of 47 tasks.
Accordingly, we designed two main test sets to evaluate both the inference and modelling capabilities of LLMs, specifically assessing their ability to handle the CQs (see \ref{sec:onto}). Through manual selection, we obtain 13 test cases, i.e. 9 inference tests and 4 modelling tests to cover the competency-question families discussed in Section~\ref{sec:onto}. In two cases, we augment the tasks to add information on external events and a planning process, which are present in the CQs but not contemplated in the original dataset.
The experiments are performed on GPT-4o with the OpenAI API, always in two configurations, that is, (i) without the ontology in the prompt and (ii)  with the ontology in the prompt, thus following the LAG approach proposed by~\cite{Gangemi2025}. The experimental dataset, tests, and code are available on GitHub\footnote{\url{https://github.com/fossr-project/ontology/tree/main/code}}.

\paragraph{\bf Inference abilities.}
The first test concerns inference accuracy, which evaluates the model's ability to detect logical inconsistencies, contradictions, and perform reasoning about agent mental state with respect to a gold standard.

In roughly a quarter of the inference subtests (2 out of 9), the ontology-augmented approach uncovered inconsistencies that the plain model overlooked. In the other cases, the two approaches led to the same conclusions. We therefore interpret the contribution of the ontology as selective: in this small sample, BDI-grounded LAG sometimes helped the model make contradictions explicit, though it did not systematically improve all cases. Below, we show two examples of the task and the results for the Location Contradiction Detection subtest (cf. Table~\ref{tab:task-list}), in which the prompt lists three everyday to-do items and asks the system whether each task–location pair is logically sound.

\begin{table}[ht]
\centering
\caption{Location Contradiction Detection subtest data.}
\label{tab:task-list}
\resizebox{.48\textwidth}{!}{
\begin{tabular}{lccc}
\toprule

\textbf{Task ID} & \textbf{Task Description}       & \textbf{Location} & \textbf{Time}       \\
\midrule
12693542         & Check into hotel                & Home              & WE-morning          \\
10169250         & Buy new water filters           & Work              & WD-evening          \\
5030401          & Workflow meeting                & Public            & WD-afternoon        \\
\bottomrule
\end{tabular}
}
\end{table}

We then prompted the LLM both with and without the ontology. Both outputs correctly identified that checking into a hotel while at home is impossible, but the ontology-augmented LLM additionally detected that buying water filters while at work also represents a contradiction. The following is an excerpt of the triples generated by the LLM when prompted with the ontology and the ultimate goal of performing the right task for the given location:

{
\small
\begin{verbatim}
@prefix bdi: <https://w3id.org/fossr/ontology/bdi/> .
@prefix ex:  <https://example.org/bdi-case#> .
@prefix rdfs:  <http://www.w3.org/2000/01/rdf-schema#> .
@prefix owl:  <http://www.w3.org/2002/07/owl#> .

ex:Agent_A1 a bdi:Agent ;
    rdfs:label "Agent reasoning about hotel check-in"@en ;
    bdi:hasBelief ex:Belief_B1 ;
    bdi:hasDesire ex:Desire_D1 ;
    bdi:hasIntention ex:Intention_I3 .

ex:Belief_B1 a bdi:Belief ;
    bdi:refersTo ex:WorldState_WS_home ;
    rdfs:comment """Agent believes they are currently at home 
        during weekend morning"""@en ;
    bdi:hasValidity ex:Interval_WE_morning .

ex:Desire_D1 a bdi:Desire ;
    bdi:isMotivatedBy ex:Belief_B1 ;
    rdfs:comment """Agent desires to perform a task appropriate to 
        the current location"""@en ;
    bdi:hasValidity ex:Interval_WE_morning .

ex:Intention_I3 a bdi:Intention ;
    bdi:fulfills ex:Desire_D1 ;
    bdi:specifies ex:Plan_P1 ;
    bdi:isJustifiedBy ex:Justification_J1 ;
    rdfs:comment """Intention: 'I do not check into the hotel while 
        I am at home'"""@en ;
    bdi:hasValidity ex:Interval_WE_morning .

ex:Plan_P1 a bdi:Plan ;
    rdfs:label "Plan to do the right task for the location"@en ;
    bdi:isSpecifiedBy ex:Intention_I3 ;
    bdi:addresses ex:Goal_G2 ;
    bdi:hasComponent ex:Task_homeActivity ;
    rdfs:comment """Plan ensuring contextual appropriateness: 
        perform home-related task instead of hotel check-in"""@en .

ex:Task_homeActivity a bdi:Task ;
    rdfs:label "perform home activity"@en ;
    bdi:occursAt ex:Location_home ;
    bdi:requiresWorldState ex:WorldState_WS_home .

ex:Goal_G2 a bdi:Goal ;
    rdfs:label "Do the right task for the location"@en ;
    owl:differentFrom ex:Goal_G1 .

ex:Goal_G1 a bdi:Goal ;
    rdfs:label "Check into the hotel"@en .

ex:Justification_J1 a bdi:Justification ;
    rdfs:label """Justification: agent cannot check into hotel 
        because of current location"""@en ;
    bdi:justifies ex:Intention_I3 ;
    rdfs:comment """
        The agent cannot check into the hotel because the 
        precondition of the task 'check into hotel' 
        (being at the hotel) contradicts the current belief
        'being at home'. The justification grounds the intention 
        not to execute the inconsistent task and instead 
        pursue a contextually valid goal.
    """@en .

ex:WorldState_WS_home a bdi:WorldState ;
    rdfs:comment "Agent is at home"@en ;
    bdi:hasLocation ex:Location_home ;
    bdi:atTime ex:Time_WE_morning .
\end{verbatim}
}

The triples describe the BDI reasoning cycle in which an agent maintains rational coherence when confronted with an inconsistent action. The agent (\texttt{ex:Agent\_A1}) believes that it is at home (\texttt{ex:Belief\_B1} referring to \texttt{ex:WorldState\_WS\_home}) and therefore forms the desire (\texttt{ex:Desire\_D1}) to act appropriately within that context. A potential action, i.e. checking into a hotel, would violate the precondition of being at the hotel, producing a contradiction between the agent’s belief and the required world state. This inconsistency gives rise to a justification (\texttt{ex:Justification\_J1}) stating that the agent cannot check into the hotel because its belief (``being at home'') conflicts with the task's precondition (``being at the hotel''). The justification supports the formation of a corrective intention (\texttt{ex:Intention\_I3}), expressed as ``I do not check into the hotel while I am at home'', which fulfils the desire to behave consistently. This intention specifies a contextually appropriate plan (\texttt{ex:Plan\_P1}) addressing the goal ``Do the right task for the location'' (i.e. \texttt{ex:Goal\_G2}) operationalised by performing a home-appropriate activity (\texttt{ex:Task\_homeActivity}). 
This example illustrates that, when the ontology is included in the prompt, the model can represent the contradiction, the inhibiting intention, and the alternative course of action using explicit BDI entities and relations. The value of the ontology here is not that it resolves the contradiction by itself, but that it provides a vocabulary and structure that make the model's reasoning more inspectable. The example also aligns with several of the CQ families listed in Table~\ref{tab:cqs-bdi}, although this should be understood as an illustrative alignment rather than a complete validation of all CQs.

\paragraph{\bf Modeling abilities.}
The second test concerns ontology-grounded modelling, which instead shows if the answer explicitly grounds its analysis in the ontology, so if it generates appropriate RDF triples. We also check if it answers the CQs.

In the modelling tests, the ontology-grounded configuration produced plausible triples aligned with the ontology and capable of representing the target mental-state transitions.

In this scenario, the agent receives a push-notification from their banking app at 10:15 AM, stating that ``Ghadeh has requested \$250 via Zelle''. Upon receiving this external event, the agent begins deliberating whether to fulfil the request and how to do so. This deliberation leads to the formation of an intention to send the payment, which is subsequently recorded as a candidate task in the agent’s plan manager: ``Task ID 11439167: \textit{pay Ghadeh} (suggested for weekday afternoon)''.

The BDI ontology allows us to explicitly trace this causal and cognitive path from event to mental process to intention to plan through formalisation in Turtle syntax.

{
\small 
\begin{verbatim}
@prefix bdi:  <https://w3id.org/fossr/ontology/bdi/> .
@prefix ex:   <http://example.org/bdi-demo/> .
@prefix xsd:  <http://www.w3.org/2001/XMLSchema#> .
@prefix rdfs:  <http://www.w3.org/2000/01/rdf-schema#> .
@prefix owl:  <http://www.w3.org/2002/07/owl#> .

ex:WorldState_WS_request a bdi:WorldState ;
    rdfs:comment """Push-notification:
        'Ghadeh has requested $250 via Zelle'."""@en ;
    bdi:atTime ex:T_2025_10_27T10_15 ;
    bdi:triggers ex:Belief_process .

ex:Agent_A a bdi:Agent .

ex:Belief_process a bdi:BeliefProcess ;
    bdi:generates ex:Belief_B ;
    bdi:isProcessedBy ex:Agent_A .  

ex:Belief_B a bdi:Belief ;
    rdfs:label "Ghadeh requested $250 via Zelle"@en ;
    bdi:refersTo ex:WorldState_WS_request ;
    bdi:motivates ex:Desire_B ;
    bdi:supports ex:Intention_B ;
    bdi:triggers ex:Desire_process ;
    bdi:isBeliefOf ex:Agent_A .  
    
ex:Desire_process a bdi:DesireProcess ;
    bdi:generates ex:Desire_B ;
    bdi:isProcessedBy ex:Agent_A .

ex:Desire_B a bdi:Desire ;
    rdfs:label "I desire to pay Ghadeh"@en ;
    bdi:refersTo ex:WorldState_WS_request ;
    bdi:triggers ex:Intention_process ;
    bdi:isDesireOf ex:Agent_A . 

ex:Intention_process a bdi:IntentionProcess ;
    bdi:generates ex:Intention_B ;
    bdi:isProcessedBy ex:Agent_A .

ex:Intention_B a bdi:Intention ;
    rdfs:label "Paying Ghadeh"@en ;
    bdi:refersTo ex:WorldState_WS_request ;
    bdi:triggers ex:Intention_process ;
    bdi:isIntentionOf ex:Agent_A ;
    bdi:fulfils ex:Desire_B .

\end{verbatim}
}

The generated triples provide a plausible formalisation of the transition from an external event to belief, desire, and intention, and they touch the CQ family concerning the dynamics of mental states (CQs 6--10).
 The world state \texttt{ex:WorldState\_WS\_request} represents the environmental fact that ``Ghadeh has requested \$250 via Zelle'', temporally anchored at \texttt{ex:T\_2025\_10\_27T10\_15}. This event triggers a belief-formation process (\texttt{ex:Belief\_process}), which the agent (\texttt{ex:Agent\_A}) processes to generate a new belief (\texttt{ex:Belief\_B}). This pattern provides direct evidence for CQ8 (i.e. Which mental process generated a given belief, desire, or intention?) and, through the process timestamp, for CQ9 (i.e. When was a mental entity generated?). The belief itself encapsulates the propositional content of the notification (``Ghadeh requested \$250 via Zelle''), refers to the world state, and motivates the subsequent desire \texttt{ex:Desire\_B}, thereby answering CQ6 (i.e. What beliefs motivated the formation of a given desire?). The belief further supports the formation of the intention \texttt{ex:Intention\_B}, reflecting epistemic continuity between cognitive layers. A dedicated desire-formation process (\texttt{ex:Desire\_process}) is triggered by the belief and generates the desire (\texttt{I desire to pay Ghadeh}), reinforcing the causal link required by CQ10 (i.e. What triggered a mental process?). This desire in turn triggers an intention-formation process (\texttt{ex:Intention\_process}) that generates the intention \texttt{ex:Intention\_B} (``Paying Ghadeh''), which fulfils the desire that motivated it, thus addressing CQ7 (i.e. Which desire does a particular intention fulfil?). Collectively, these triples instantiate a temporally ordered cognitive workflow explicitly annotated with triggering relations, generation events, and fulfilment links that operationalise the BDI ontology's competency questions CQ6–CQ10 within a single, coherent reasoning chain.

Overall, the LAG experiment should be read as a proof of feasibility. The sample is small, the assessment is partly qualitative, and the comparison does not fully isolate the effect of the ontology from the effect of prompt design. What the experiment supports is however that when the BDI ontology is included in the prompt within a LAG setting, the LLM can, in some cases, produce more explicit and auditable representations of contradictions and mental-state dynamics. At the same time, the current BDI ontology does not explicitly axiomatise intention--intention conflict; such contradictions emerge only when incompatible intentions are grounded in mutually exclusive beliefs or goals and are surfaced by contextual reasoning rather than by ontology axioms alone.

 This design choice is deliberate, as it keeps the core vocabulary flexible, in that agents are allowed to possess mutually incompatible intentions before committing, and it is the reasoner (or a LAG layer \cite{Gangemi2025}) that surfaces the contradiction basing on the context, not the ontology axioms themselves. 

\subsection{Using the ontology into BDI frameworks}
\label{sec:semas}

The BDI ontology can be effectively integrated into real-world scenarios involving executable BDI frameworks. In particular, we consider \textsc{Semas}\footnote{\url{https://github.com/cfabiolongo/Semas}}, a Prolog-style reasoning platform that enables logic-based agents. 

This integration establishes a unified conceptual and operational bridge between symbolic mental-state representations and logic-based procedural reasoning.
 The BDI ontology offers a semantic model for describing agents’ mental states and their interrelations, \textsc{Semas} endows these entities with operational semantics, enabling agents to reason over them and act upon them at runtime within a rule-based environment interoperable with the Semantic Web.

In \textsc{Semas}, reasoning processes are implemented through production rules of the form

\begin{center}
    \texttt{[HEAD] / [CONDITIONALS] >> [TAIL]},
\end{center}

\noindent 
where the \texttt{[HEAD]} denotes the triggering belief, the \texttt{[CONDITIONALS]} define contextual constraints or auxiliary beliefs required for activation, and the \texttt{[TAIL]} specifies the list of actions to be executed when the rule fires. 
This structure naturally maps onto the BDI ontology. A belief in the BDI ontology corresponds to a declarative assertion in the \textsc{Semas} Knowledge Base (KB) that can appear either as a \texttt{[HEAD]}(triggering rule execution) or as part of \texttt{[CONDITIONALS]} (contextual preconditions).
Desires and intentions, expressed in RDF as goal-oriented entities (e.g., \texttt{:Goal}, \texttt{:Desire}, \texttt{:Intention}), are operationalised in \textsc{Semas} as executable rule plans, where the \texttt{[TAIL]} encodes the actions that realise such intentions.

The ontology therefore provides the semantic layer, defining relations such as \texttt{:motivates}, \texttt{:fulfils}, and \texttt{:triggers},
while \textsc{Semas} provides the procedural substrate that enacts these relationships through executable rules.
For instance, when a world state in RDF triggers a belief-formation process (i.e. \texttt{bdi:BeliefProcess}), \textsc{Semas} evaluates a corresponding rule whose \texttt{[HEAD]} matches that belief, checks the required \texttt{[CONDITIONALS]} (e.g. verification of trust or temporal validity), and then executes the \texttt{[TAIL]} (e.g. forming a new desire or plan). 
The integration allows RDF-described mental entities to be directly queried and manipulated via \textsc{Semas}' Prolog-style inference engine, enabling hybrid reasoning where symbolic knowledge from the Semantic Web dynamically informs rule execution.

In essence, the BDI ontology defines what an agent's mental architecture is, while a BDI framework, such as \textsc{Semas}  defines how this architecture operates. Together, they establish a semantically grounded, executable BDI framework in which the ontology provides the declarative semantics of agency, and the operational component provides the procedural logic that brings those semantics to life.

This research direction has been presented in ~\cite{Longo2025}, including the integration of a very preliminary version of BDI ontology further extended in this work, with \textsc{Semas} through the Triples-to-Beliefs-to-Triples (T2B2T) paradigm.

This paradigm establishes a seamless bridge between agents' mental attitudes (Beliefs, Desires, Intentions) and RDF triples describing their domain.
 In this approach, domain knowledge encoded as triples is dynamically translated into beliefs within the agent's internal knowledge base, where reasoning unfolds according to the BDI model. The outcomes of this reasoning are then projected back into RDF, completing a feedback loop that keeps the agent's cognitive state aligned with the external knowledge graph. 
This integration enables multi-agent systems to operate coherently within Semantic Web ecosystems, where agents can perceive, reason, and act using the same ontological vocabulary that structures the domain. The result is a unified architecture in which the BDI ontology supplies the conceptual semantics of agency, while \textsc{Semas} and the T2B2T paradigm provide the executable mechanism linking symbolic reasoning with the Web of Data.

\paragraph{\bf Case Study Revisited through the BDI Ontology}

In pratical as reported in \cite{Longo2025} as a case study, a Knowledge Graph (KG) describing Italian scholars in 2024 by extracting data from the SCOPUS and processing identified co-authorship relations and top-authors were translated into RDF triples and stored in a GraphDB triple store, producing a KG of approximately 100,000 triples.

Within the SEMAS framework, these RDF triples are interpreted through a first version of BDI ontology, where co-authorship and top-authors are mapped into \textit{beliefs}. In particular, information such as co-authorships and top-authorships is represented as instances of the \texttt{DerivedBelief} class, enabling agents to reason over structured knowledge extracted from the KG.

The triples-to-beliefs (T2B) pipeline operationalizes this mapping: SPARQL queries retrieve relevant triples (e.g., \texttt{isTopAuthorIn}, \texttt{coAuthorWith}, \texttt{hasAffiliationWith}) and assert corresponding beliefs in the agents’ knowledge base. These beliefs constitute the informational state of the agents in accordance with the BDI paradigm.

On top of this belief layer, SEMAS production rules implemented in Phidias define the agents’ \textit{intentions}. These intentions correspond to procedural plans activated when specific belief conditions hold. For instance, when a scholar is identified as a top-author and relevant co-authorship relations are detected, inference rules trigger plans aimed at identifying advantageous institutional affiliations. Such plans operationalize decision-making strategies inspired by small-world network theories.

The execution of these intentions is driven by explicit or implicit \textit{goals}, which represent desired states of the world that agents aim to achieve. Goals are thus achieved through the execution of rule plans, which may involve asserting new beliefs or retracting outdated ones (e.g., updating affiliations or selection statuses).

At the end of the SEMAS reasoning process, the system produces:
(i) updated beliefs reflecting newly inferred knowledge (e.g., potential affiliations and collaborations), and
(ii) achieved goals, resulting from the successful execution of plans encoded in the agents’ intentions.

\section{Conclusions and future work}
\label{sec:conclusions}
This paper introduced a BDI ontology that provides a structured and temporally grounded representation of agents’ mental states and their dynamics. In contrast to more lightweight or static representations, the proposed ontology explicitly models (i) the causal and procedural relations between beliefs, desires, and intentions, (ii) the role of mental processes in generating and updating such states, and (iii) their temporal validity, enabling the reconstruction of an agent’s cognitive state over time.

The adoption of the eXtreme Design methodology ensured a systematic alignment between user stories, competency questions, and the resulting conceptual model, supporting traceability from requirements to formalisation.

With respect to applications, we illustrated how the ontology can be integrated within an executable BDI framework (SEMAS), where it provides a semantic layer grounding rule-based reasoning in explicit mental-state representations. In addition, we explored its use in conjunction with large language models in a LAG setting. The results suggest that, when included in the prompt, the ontology can support the generation of more explicit and inspectable representations of mental-state dynamics and potential inconsistencies. However, these findings should be interpreted as preliminary, as the current evaluation does not fully isolate the contribution of the ontology from that of prompt design.

Overall, the contribution of this work lies in providing a semantically grounded, process-oriented, and temporally explicit model of BDI reasoning that can serve as a foundation for interoperable, explainable, and hybrid symbolic–neural agent systems.

\section*{Generative AI Statement}
During the preparation of this work, the authors used GPT-4 to assist with grammar and spelling revision and to generate draft content for the online documentation hosted in the project's GitHub repository. After using this tool, the authors carefully reviewed and edited all generated content as appropriate and take full responsibility for the content of this publication.

\section*{Acknowledgements}
This work was supported by the FOSSR (Fostering Open Science in Social Science Research) project, funded by the European Union - NextGenerationEU under NRRP Grant agreement n. MUR IR0000008.

\bibliographystyle{unsrt}
\bibliography{references}
\newpage

\appendix

\section {User stories}
\label{sec:app}
Ontological requirements were collected from stakeholders in the form of user stories, which were subsequently decomposed to derive the corresponding competency questions (CQs). We drew inspiration from \cite{presutti2009extreme} and \cite{de2023polifonia} for the formalization of user stories, supporting both the elicitation of user requirements and their refinement throughout the different phases of the iterative development process that led to the definition of the ontology.

\subsection{Requirement story card: World, Agents and Mental Entities}
\paragraph{\bf User story.} 
Mark is a student enrolled in a Computer Science degree programme. He is aware of the structure of his programme, including the courses he must take, the academic calendar, and the schedule of examinations. 

Based on his experience and understanding, Mark believes that graduates in Computer Science have good employment prospects. He also considers himself strong in scientific subjects, enjoys working with computers for long periods, and feels confident in his logical reasoning skills. Because of this, Mark aims to complete his degree. 

To achieve this, he focuses on several key steps, such as passing his exams, preparing a thesis, and eventually graduating. Mark organises his activities accordingly. For example, he plans when to take exams and how to prepare for them, breaking down his study into manageable tasks over time.

\paragraph{\bf Story decomposition.} The agent has, i.e. 
\begin{itemize}
\item subjective knowledge, based on his beliefs on (i) graduates' employment prospects, (ii) his aptitude for science, (iii) his ability to work long hours at the computer, and (iv) his aptitude for logical reasoning.
\item subjective knowledge based on his beliefs on (i) graduates' employment prospects, (ii) his aptitude for science, (iii) his ability to work long hours at the computer, and (iv) his aptitude for logical reasoning.
\item goals and intentions centred on completing a degree programme, passing exams, preparing a thesis, and graduating.
\end{itemize}

\paragraph{\bf Extracted CQs}
\begin{itemize}
\item What are mental entities? $\rightarrow$ \textbf{CQ1}
\item What mental states does an agent hold? $\rightarrow$ \textbf{CQ2}
\item What are the constituent mental entities? $\rightarrow$ \textbf{CQ3}
\item What mental processes has an agent undergone? $\rightarrow$ \textbf{CQ4}
\item What is the world state of a mental state? $\rightarrow$ \textbf{CQ5}
\item What is the temporal validity of a mental state? $\rightarrow$ \textbf{CQ16}
\item What mental states were valid at a given time? $\rightarrow$ \textbf{CQ18}
\end{itemize}

\subsection{Requirement story card: Dynamics of Mental States}
\paragraph{\bf User story.} 
Mark is a student pursuing a degree in Computer Science and is steadily progressing through his academic programme. Over time, he has developed a clear understanding of his progress and believes that he has successfully passed all required examinations so far. Based on this belief, he continues planning his studies with the expectation of moving on to more advanced courses and ultimately graduating on time.

One day, however, Mark receives an official notification stating that he has failed one of the required examinations. This new piece of information forces him to reconsider his previous understanding of his academic situation.

As a result, Mark updates his belief about his progress: he no longer believes that he has successfully completed that exam. This change has immediate consequences. Since passing all required exams is necessary for graduation, his broader objective is affected. Mark now realises that in order to graduate, he must first retake and pass the failed exam.

This leads him to revise his priorities. He develops a stronger focus on retaking the exam as soon as possible, adjusting his plans accordingly. Instead of proceeding directly to the next steps in his academic path, he now allocates time for additional study and preparation.

Mark reorganises his schedule by introducing new study sessions and planning to register for the next available exam session. His previous plan is therefore revised, and a new plan is established that reflects the updated situation.

\paragraph{\bf Story decomposition.} 
The agent has, i.e.
\begin{itemize}
    \item an initial belief about his academic progress, namely that all required examinations have been successfully passed, which supports his expectation of progressing towards graduation;

    \item an updated belief triggered by an external event (the failure notification), leading to a revision of his understanding of his academic state;

    \item a dependency between beliefs and higher-level objectives, where the revision of a belief impacts the feasibility of the goal of graduating on time;

    \item a revised desire focused on retaking the failed examination as a necessary step to achieve the overall objective;

    \item updated intentions that reflect new priorities, in particular the commitment to prepare for and register for a resit;

    \item a reconfiguration of plans, where the original plan is replaced by a new one including additional study activities and a new examination attempt;

    \item a set of mental processes (belief revision, desire update, intention formation) triggered by the external event and responsible for the evolution of the agent’s mental state;

    \item temporally bounded mental states, where the original belief remains valid until the failure is observed, after which a new belief is generated;

    \item a traceable evolution of mental entities over time, enabling reconstruction of the agent’s cognitive state at different points in time.
\end{itemize}

\paragraph{\bf Extracted CQs}
\begin{itemize}
    \item What beliefs motivated the formation of a given desire?$\rightarrow$ \textbf{CQ6}
    \item   Which desire does a particular intention fulfil?$\rightarrow$ \textbf{CQ7}
    \item  Which mental process generated a given belief, desire, or intention?$\rightarrow$ \textbf{CQ8}
    \item When was a mental entity generated?$\rightarrow$ \textbf{CQ9}
    \item What triggered a mental process?$\rightarrow$ \textbf{CQ10}
    \item What justifications support a specific mental entity?$\rightarrow$ \textbf{CQ11}
    \item  What is the temporal validity (start and end time) of a mental state?$\rightarrow$ \textbf{CQ16}
    \item What mental states were valid at a specific point in time? $\rightarrow$ \textbf{CQ17}
    \item How has a mental entity evolved over time?$\rightarrow$ CQ18
\end{itemize}

\subsection{Requirement story card: Dynamics of Mental States}
\paragraph{\bf User story.} 
Once Mark has decided what he wants to achieve, he needs to turn his intentions into concrete actions.

Mark has a clear objective: to obtain a degree in Computer Science. This objective guides all his decisions and serves as a reference point for organising his activities.

To achieve this goal, Mark develops a structured plan. He considers what needs to be done, in which order, and under which conditions. Based on his current situation and commitments, he identifies the most appropriate sequence of actions that will allow him to progress toward graduation.

This plan includes several steps, such as enrolling in the programme, attending courses, registering for exams, preparing for them, passing them, paying tuition fees, and eventually attending the graduation ceremony. These activities are not independent, but organised in a specific order, where some actions must be completed before others can take place.

Each activity is also constrained by time. For example, registering for an exam is only possible within a specific time window, and missing that window may delay the entire plan.

When unexpected events occur — such as failing an exam — Mark must revise his plan. He adapts it by introducing new actions, such as additional study sessions or registering for a resit, while maintaining his overall objective.

At any moment, it is possible to trace how Mark’s plan evolved over time, from the initial version to the updated one. Despite these changes, all actions remain aligned with his original goal, ensuring that his behaviour remains coherent and goal-directed.

\paragraph{\bf Story decomposition.} 

The agent has, i.e.
\begin{itemize}
    \item a goal representing the desired outcome, namely obtaining a Computer Science degree, which guides the agent’s overall behaviour;

    \item intentions that must be operationalised into actionable commitments, serving as the basis for defining concrete courses of action;

    \item a planning process that reasons upon current intentions and contextual constraints to generate a structured plan;

    \item a plan defined as an ordered sequence of tasks, where activities such as enrolling, attending courses, registering for exams, and graduating are organised according to precedence relations;

    \item dependencies among tasks, such that certain activities must be completed before others can be executed;

    \item temporally constrained tasks, where each action is valid only within a specific time interval (e.g. examination registration windows);

    \item a relation between intentions and plans, where intentions specify the plans that operationalise them;

    \item a relation between plans and goals, where each plan is directed towards achieving a specific goal;

    \item the capability to revise plans in response to external events, such as failing an examination, leading to the introduction of new tasks and the adaptation of the existing plan;

    \item a traceable evolution of plans over time, allowing reconstruction of how the plan changes from its initial version to subsequent revisions while preserving goal alignment.
\end{itemize}
\paragraph{\bf Extracted CQs}
\begin{itemize}
    \item What goal does a given intention or plan aim to fulfil?$\rightarrow$ \textbf{CQ12}
    \item   What plan has been specified by a particular intention?$\rightarrow$ \textbf{CQ13}
    \item   What planning process led to the creation of a particular plan? $\rightarrow$ \textbf{CQ14}
    \item What is the ordered sequence of tasks that compose a given plan? $\rightarrow$ \textbf{CQ15}
    \item  What is the temporal validity (start and end time) of a mental state?$\rightarrow$ \textbf{CQ16}
    \item How has a mental entity evolved over time?$\rightarrow$ \textbf{CQ18}
\end{itemize}

\section{Ontology formalisation}
\label{sec:app2}
\subsection{Ontology formalisation}

This appendix reports the complete formalisation of the BDI ontology, including the full set of description logic axioms underlying the conceptual model introduced in the main text.
For this, it defines key mental states and their relationships, thereby capturing the reasoning, motivation, and commitment to action of the agent. Figure~\ref{fig:schema_bdi} shows the Graffoo~\cite{falco2014modelling} diagram of the BDI ontology.  

\paragraph{\bf World, agents and mental entities.}
\begin{align}
& \textit{WorldState} \sqsubseteq \textit{d0:Eventuality}  \label{dl:core_s}\\
& \textit{WorldState} \sqsubseteq \forall \textit{isPerceivedBy.Agent} \\
& \textit{MentalEntity} \sqsubseteq \textit{d0:CognitiveEntity}\\
& \textit{MentalEntity} \sqsubseteq \forall\textit{hasPart.MentalEntity} \\
& \textit{MentalEntity} \sqsubseteq \exists\textit{refersTo.WorldState} \\
& \textit{Agent} \sqsubseteq \textit{dul:Agent} \\
& \textit{Agent} \sqsubseteq \exists \textit{perceives.WorldState} \\
& \textit{Agent} \sqsubseteq \forall \textit{cognises.MentalEntity} 
\label{dl:core_e}
\end{align}

In this setting, the class \texttt{:WorldState} represents the state of the environment or context as perceived or observed by an agent. This class provides the factual or situational grounding upon which an agent forms its beliefs and generates desires. A \texttt{:WorldState} may encapsulate conditions, events, or changes in the external world relevant to the agent's reasoning and decision-making processes. From the foundational perspective, the \texttt{:WorldState} is a subclass of \texttt{d0:Eventuality}, which is the class defined in DOLCE-0\footnote{DOLCE-0 is a supplementary ontology used as a generalisation of DOLCE+DnS Ultralite. DOLCE-0 is available online at \url{http://www.ontologydesignpatterns.org/ont/d0.owl}.} for distinguish between event occurrences, their types, and constructed objects. The object property \texttt{:isPerceivedBy} and its inverse \texttt{:perceives} ensure that a \texttt{:WorldState} is perceived by an \texttt{:Agent}. The class \texttt{:Agent} is a subclass of \texttt{dul:Agent} and represents every autonomous entity capable of perceiving its environment, reasoning about it, and acting upon it to achieve specific goals or desires. Agents cognise mental entities, which are represented by the class \texttt{:MentalEntity}, a subclass of \texttt{d0:CognitiveEntity}, which allows mental entities to be treated as a mental object or content. In this ontology, to cognise, which is represented by the object property \texttt{:cognises}, means to mentally represent or be aware of an entity. This denotes an agent's epistemic relation to mental content (e.g. beliefs, desires, and intentions). 
The ontology also introduces the property \texttt{:hasPart} to formalise meronymic relations, i.e. part–whole structures, among mental entities. This property enables the representation of complex mental constructs as composed of finer-grained elements, thus reflecting the inherently structured nature of cognition. For example, a \texttt{:MentalEntity} may itself be composed of multiple parts that contribute to its overall content. Consider an agent holding the belief {\em ``I believe the meeting will start at 10 a.m. in Room 5''}. This belief can be decomposed into at least two parts: (i) the component concerning time (i.e. the meeting will start at 10 a.m.) and (ii) the component concerning location (i.e. the meeting will be held in Room 5). By using the \texttt{:hasPart} property, the ontology can capture this internal structure, ensuring that mental entities can be modelled not only as atomic states but also as compositional entities with meaningful internal relations. Such a representation supports more fine-grained reasoning, for instance, by allowing the system to update only a part of a belief (e.g. the meeting location) while leaving the rest unchanged.

\Crefrange{dl:mentalstate_s}{dl:mentalstate_e} introduce beliefs, desires and intentions in the ontology.

\begin{align}
& \textit{MentalState} \sqsubseteq \textit{MentalEntity} \label{dl:mentalstate_s}\\
& \textit{MentalState} \sqsubseteq \forall\textit{hasPart.MentalState} \\
& \textit{Agent} \sqsubseteq \exists \textit{hasMentalState.MentalState} \\
& \textit{Belief} \sqsubseteq \textit{MentalState}  \\
& \textit{Belief} \sqsubseteq \forall\textit{hasPart.Belief} \\
& \textit{Belief} \sqsubseteq \exists\textit{motivates.Desire} \\
& \textit{Belief} \sqsubseteq \exists\textit{supports.Intention} \\
& \textit{Desire} \sqsubseteq \textit{MentalState}  \\
& \textit{Desire} \sqsubseteq \forall\textit{hasPart.Desire} \\
& \textit{Intention} \sqsubseteq \textit{MentalState} \\
& \textit{Intention} \sqsubseteq \forall\textit{hasPart.Intention} \\
& \textit{Desire} \sqsubseteq \exists \textit{isMotivatedBy.Belief} \\
& \textit{Intention} \sqsubseteq \exists \textit{fulfils.Desire} \\
& \textit{Intention} \sqsubseteq \exists \textit{isSupportedBy.Belief} \label{dl:mentalstate_e}
\end{align}

Mental states, such as \texttt{:Belief}, \texttt{:Desire}, and \texttt{:Intention}, represent the informational and motivational aspects of the cognition of an agent. More specifically, a \texttt{:Belief} represents the mental state of an agent regarding something that the agent believes to be true. It captures the subjective perception or understanding of the world by an agent, which may or may not align with objective reality. i.e. a \texttt{:WorldState}. Then, a \texttt{:Desire} represents a motivational mental state of an agent, encapsulating what the agent wishes or aspires to bring about in the world. Desires are expressions of preferences, but unlike intentions, they do not imply a commitment to act. Desires serve as the driving force behind an agent's decision-making process, often interacting with beliefs and intentions to influence behaviour. Finally, an \texttt{:Intention} represents a deliberative mental state of an agent, characterised by the agent’s commitment to achieving a specific goal or executing a plan. Unlike a desire, which expresses a motivational preference, an intention reflects a higher degree of resolve, where the agent actively decides to pursue the desired outcome.
Intentions bridge the gap between an agent's desires and actions, driving goal-oriented behaviour based on the agent's beliefs about feasibility and circumstances.
Intentions depend on beliefs about feasibility and current conditions (e.g. {\em ``I believe the store is open.''}).
Intentions emerge from prioritised desires or goals (e.g. {\em ``I desire to buy groceries.''}).

\paragraph{\bf Dynamics of mental states.} Mental states can influence and be influenced by mental processes, which govern their generation, modification, and suppression over time. In the ontology the core for representing the dynamics of mental states is formalised by \crefrange{dl:process_s}{dl:process_e}.

\begin{align}
& \textit{MentalProcess} \sqsubseteq \textit{d0:Activity} \label{dl:process_s}\\
& \textit{MentalProcess} \sqsubseteq \textit{MentalEntity}  \\
& \textit{MentalProcess} \sqsubseteq \forall\textit{hasPart.MentalProcess} \\
& \textit{MentalProcess} \sqsubseteq \exists \textit{isProcessedBy.Agent} \\
& \textit{MentalProcess} \sqsubseteq \exists\textit{reasonsUpon.MentalState}  \\
& \textit{MentalProcess} \sqsubseteq \forall\textit{isTriggeredBy.MentalEntity}  \\
& \textit{MentalProcess} \sqsubseteq \exists\textit{affects.MentalState}  \\
& \textit{generates} \sqsubseteq \textit{affects}  \\
& \textit{modifies} \sqsubseteq \textit{affects}  \\
& \textit{suppresses} \sqsubseteq \textit{affects} \label{dl:process_e} 
\end{align}

The class \texttt{:MentalProcess} is defined as a subclass of both \texttt{:MentalEntity} and \texttt{d0:Activity}. As such, a mental process is a specific type of mental entity that represents an action or task processed by an agent. Its execution is grounded in one or more existing and valid mental states, which are linked to the process via the object property \texttt{:reasonsUpon}. A mental process can be triggered by any mental entity. Consequently, the dynamics of mental states can be activated either by another mental state or by a mental process. For example, a belief such as {\em ``I believe the store is open''} may trigger a mental process that affects a corresponding desire, such as {\em ``I desire to buy groceries''}. The relation that allows a mental process to affect a mental state is expressed in the ontology by the object property \texttt{:affects}. The inverse relation is enabled by the corresponding inverse object property \texttt{:isAffectedBy} that accepts a mental state and a mental process as domain and range, respectively. We distinguish three ways in which a mental process can affect a mental state. The first, \texttt{:generates}, has a generative nature and is applied when a mental process generates a new mental state. The second \texttt{:modifies} is materialised when a mental process modifies an existing mental state. Finally, the third, \texttt{:suppresses}, is materialised when a mental process results in the suppression of a mental state.
With respect to mental state processing, we distinguish three types of processes: (i) \texttt{:BeliefProcessing}, (ii) \texttt{:DesireProcessing}, and (iii) \texttt{:IntentionProcessing}, as formalised by \crefrange{dl:process_special_s}{dl:process_special_e}.

\begin{align}
& \textit{BeliefProcess} \sqsubseteq \textit{MentalProcess} \label{dl:process_special_s}\\
& \textit{BeliefProcess} \sqsubseteq \forall\textit{affects.Belief} \\
& \textit{DesireProcess} \sqsubseteq \textit{MentalProcess} \\
& \textit{DesireProcess} \sqsubseteq \forall\textit{affects.Desire} \\
& \textit{IntentionProcess} \sqsubseteq \textit{MentalProcess} \\ 
& \textit{IntentionProcess} \sqsubseteq \forall\textit{affects.Intention} \label{dl:process_special_e} 
\end{align}

The class \texttt{:BeliefProcess} represents the cognitive process through which an agent generates, modifies, or suppresses beliefs based on a given contextual setting comprising its mental states, perception of the world, and ongoing mental processing.
This process enables agents to maintain a dynamically evolving mental representation of both their environment and internal states. For example, an agent monitoring a public health database may initially believe that a disease outbreak is contained. However, upon processing new epidemiological reports, the agent updates its belief to reflect the increased risk of a wider spread and adjusts its reasoning accordingly.

Instead, the class \texttt{:DesireProcess} represents the mental process by which an agent generates, modifies, or suppresses desires based on internal motivations derived from its mental states or contextual changes in the environment.
This process helps agents structure goal-oriented behaviour before intentions are formed. For example, a decision-support agent in a climate policy system may initially lack any specific desire regarding renewable energy subsidies. After analysing recent economic and environmental reports, it generates a new desire to advocate for increased subsidies, which may subsequently influence its intentions and actions.

Finally, the class \texttt{:IntentionProcess} represents the process by which an agent selects, commits to, and refines intentions based on its desires, available resources, and deliberative reasoning. It determines which desires are pursued as actionable commitments and adapts those commitments as new information becomes available. For example, an autonomous assistant tasked with monitoring policy compliance may develop a desire to verify the implementation of a new regulation. After assessing feasibility and priority, it forms an intention to generate a compliance report, refining that intention further as additional data becomes available.

Mental processes allow the materialisation of direct relations among mental states. These relations are formalised in \crefrange{dl:mental_state_rels_s}{dl:mental_state_rels_e}.

\begin{align}
& \textit{Belief} \sqsubseteq \exists\textit{motivates.Desire} \label{dl:mental_state_rels_s}\\
& \textit{Belief} \sqsubseteq \exists\textit{supports.Intention}  \\
& \textit{Intention} \sqsubseteq \exists\textit{fulfils.Desire}  
\label{dl:mental_state_rels_e} 
\end{align}

Beliefs motivate and support the formation of desires and intentions, respectively, while intentions serve to fulfils desires. These relationships among mental states can give rise to structured chains of cognitive dependencies that reflect how an agent reasons over time. By tracing the progression from a belief to a desire, and from that desire to an intention, it becomes possible to reconstruct the agent's deliberative process. These cognitive chains are particularly valuable for enabling explainability, as they allow external observers or systems to understand the underlying rationale for an agent's actions.
Importantly, such reasoning paths can incorporate both temporal constraints (cf. the paragraph on temporal reasoning below) and environmental conditions. Temporal information helps determine when a mental state, such as a belief, was formed or revised, thereby providing insight into the evolving nature of the agent’s internal state. Environmental conditions, on the other hand, make it possible to identify which contextual changes or external observations influenced the emergence or transformation of mental states. As a result, this modelling approach not only supports the simulation of intelligent behaviour but also enhances the transparency and contextual grounding of agent-based explanations. 

Furthermore, the class \texttt{:Justification} is introduced to provide explicit support for explainability within the ontology as reported by \crefrange{dl:mental_state_rels_s}{dl:mental_state_rels_e}. 

\begin{align}
& \textit{Justification} \sqsubseteq \textit{dul:Description} \label{dl:justification_s}\\
& \textit{Justification} \sqsubseteq \exists\textit{justifies.MentalEntity}  
\label{dl:justification_e} 
\end{align}

Modelled as a subclass of \texttt{dul:Description}, a justification captures the reason, evidence, or rationale underlying the existence of a particular mental entity. This design makes it possible to link beliefs, desires, or intentions to the contextual or inferential basis from which they arise. For example, a belief such as {\em``the meeting will be postponed''} may be associated with a justification referring to the agent's observation of an official announcement. By representing justifications as first-class entities, the ontology enables the reconstruction of the deliberative trace behind mental states, thus supporting transparency and interpretability in agent reasoning. This modelling choice ensures that the ontology does not merely encode mental entities as isolated constructs, but also provides the explanatory grounding needed for advanced explainability and trust in cognitive systems.

\paragraph{\bf Goals and planning.}
The ontology models how an agent's mental states, i.e. beliefs, desires, and intentions, progressively contribute to the realisation of goal-directed behaviour. At the heart of this cognitive architecture lies the principle that the ultimate purpose of these mental states is to support the elaboration of a plan aimed at achieving a given goal. Among these states, intentions play a pivotal role: they represent the actionable commitments of the agent and serve as cognitive input to a planning process. \Crefrange{dl:goals_s}{dl:goals_e} formalises the part of the ontology focused on goals and planning.

\begin{align}
& \textit{Plan} \sqsubseteq \textit{dul:Plan} \label{dl:goals_s}\\
& \textit{Goal} \sqsubseteq \textit{dul:Goal}  \\
& \textit{Intention} \sqsubseteq \exists\textit{specifies.Plan}   \\
& \textit{Plan} \sqsubseteq \exists\textit{addresses.Goal}  \\
& \textit{Planning} \sqsubseteq \textit{MentalProcess}  \\
& \textit{Planning} \sqsubseteq \exists\textit{defines.Plan}  \\
& \textit{Planning} \sqsubseteq \forall\textit{reasonsUpon.Intention}  \\
& \textit{Planning} \sqsubseteq \forall\textit{hasPart.Planning}  \\
& \textit{Planning} \sqsubseteq \exists\textit{reasonsUpon.Intention}  
\label{dl:goals_e} 
\end{align}

This planning process is conceptualised as a mental process in its own right. This takes into account the intentions (i.e. restrictions of the object property \texttt{:reasonsUpon}), available resources, and contextual constraints of the agent to generate a plan, which is aligned with DOLCE by asserting that \texttt{:Plan} is a subclass of \texttt{dul:Plan}. This ensures ontological compatibility with standard descriptions of plans as complex descriptions or specifications of intended behaviour. The resulting plan is a structured sequence of actions or operations explicitly designed to fulfil a specific goal, which is represented by the class \texttt{:Goal} modelled a subclass of \texttt{dul:Goal}.  By modelling these relationships, the ontology enables representation and reasoning over how plans address goals, how intentions initiate planning, and how planning processes result in concrete strategies for action. For example, an intelligent agent responsible for disaster response may have the intention of delivering medical supplies to a remote area. Through the Planning process, it determines the optimal route, identifies necessary resources, and structures the sequence of actions required to execute the delivery efficiently.
In our ontology, goals are not modelled as a special kind of desire (i.e. as subclasses of \texttt{:MentalState}), as is sometimes done in BDI-inspired systems (e.g. Braubach et al.~\cite{Braubach2004} define goals as motivational attitudes derived from desires). Instead, we treat goals as descriptions, that is, intentional objects of planning, and place them in the class \texttt{dul:Goal}. By modelling goals as descriptions rather than mental states, we maintain a clean separation between the cognitive layer (beliefs, desires, intentions) and the actionable planning layer (goals and plans). This approach also allows goals to be shared, agreed upon, or communicated across multiple agents without conflating them with any particular agent's motivational dynamics. In doing so, our design is consistent with the position of Winikoff et al. (2002), who argue that goals have two aspects: (i) a declarative aspect, understood as a description of the state of affairs being sought, and (ii) a procedural aspect, corresponding to the set of plans available for achieving the goal. In such a design, agents can attribute the same goal description to themselves or others, and can build distinct but aligned plans based on that shared goal. The advantage is that reasoning about goals remains distinct from reasoning about agent-specific mental attitudes, thereby enabling modularity, explainability, and interoperability in multi-agent settings.

Plans can be described as structured sequences of actions following the schema provided by the Sequence pattern\footnote{\url{http://ontologydesignpatterns.org/index.php/Submissions:Sequence}} as formalised by \crefrange{dl:actions_s}{dl:actions_e}. Each action within a plan is represented by an instance of the class \texttt{:Action\-Description}, which captures the descriptive dimension of an intended action. A \texttt{:Plan} always begins and ends with some  \texttt{:Task}. This is enabled by the restrictions on the object properties \texttt{:beginsWith} and \texttt{:endsWith} associated with the class \texttt{:Plan}. A \texttt{:Task} may follow or precede another \texttt{:Task} through the transitive object properties \texttt{:follows} and \texttt{:precedes}. Additionally, plans can be structured meronymically through the \texttt{:hasPart} relation, enabling the decomposition of complex plans into constituent tasks or sub-plans. By modelling plans and tasks in this way, the ontology supports fine-grained specification and reasoning over the internal structure of plans, enabling the representation of ordered sequences of tasks and their dependencies.

\begin{align}
& \textit{Task} \sqsubseteq \textit{dul:Task} \label{dl:actions_s}\\
& \textit{Task} \sqsubseteq \forall\textit{follows.Task} \\
& \textit{Plan} \sqsubseteq \exists\textit{hasComponent.Task} \\  
& \textit{Plan} \sqsubseteq \exists\textit{beginsWith.Task} \\
& \textit{Plan} \sqsubseteq \exists\textit{endsWith.Task} \\
& \textit{Plan} \sqsubseteq \exists\textit{hasPart.Plan}
\label{dl:actions_e} 
\end{align}

While a \texttt{:Plan} provides a structured description of actions
intended to achieve a \texttt{:Goal}, a \texttt{:PlanExecution} captures the situated real-world unfolding of that plan. This includes the execution of actions, interaction with the environment, and the potential influence of contingencies or external events. Hence, the class \texttt{:PlanExecution} represents the actual enactment of a \texttt{:Plan} by an \texttt{:Agent} within a specific temporal and environmental context. \Crefrange{dl:planning_s}{dl:planning_e} formalises the part of the ontology focused on plan execution.

\begin{align}
& \textit{PlanExecution} \sqsubseteq \textit{dul:PlanExecution} 
\label{dl:planning_s}\\
& \textit{PlanExecution} \sqsubseteq \exists\textit{satisfies.Plan}   \\
& \textit{PlanExecution} \sqsubseteq \exists\textit{addresses.Goal}  \\
& \textit{PlanExecution} \sqsubseteq \exists\textit{hasComponent.Action} \\ 
& \textit{PlanExecution} \sqsubseteq \exists\textit{isExecutedBy.Agent} \\ 
& \textit{PlanExecution} \sqsubseteq \exists\textit{bringsAbout.WorldState}\\
& \textit{PlanExecution} \sqsubseteq \exists\textit{atTime.TemporalEntity}\\
& \textit{Action} \sqsubseteq \textit{dul:Action} \\
& \textit{Action} \sqsubseteq \exists\textit{isExecutionOf.Task} \\
& \textit{Action} \sqsubseteq \exists\textit{isPerformedBy.Agent} \\
& \textit{Action} \sqsubseteq \exists\textit{bringsAbout.WorldState}\\
& \textit{Action} \sqsubseteq \exists\textit{atTime.TemporalEntity}
\label{dl:planning_e} 
\end{align}

The formalisation of planning and its execution is captured through a set of axioms that distinguish between the cognitive process of planning and the operational realisation of plans. The class \texttt{:PlanExecution} is defined as a subclass of \texttt{dul:PlanExecution}, thereby aligning it with the DOLCE Ultra Lite foundational ontology and situating it within the broader category of event executions. Each instance of \texttt{:PlanExecution} is linked to a \texttt{:Plan} it satisfies (i.e. the existential restriction on the property \texttt{:satisfies}) and to the \texttt{:Goal} it addresses (i.e. the existential restriction on the property \texttt{:addresses}), enabling traceability from high-level objectives to concrete actions carried out in the environment. A plan execution is composed of one or more \texttt{:Action} instances (i.e. the existential restriction on the property \texttt{:hasComponent}), reflecting the operational realisation of the plan's structure. The agent responsible for carrying out the plan is linked through the \texttt{:isExecutedBy} property.
Furthermore, plan executions are associated with a specific \texttt{:WorldState} they bring about (i.e. the existential restriction on the property \texttt{:bringsAbout}), supporting causal reasoning over their outcomes. Each execution is also temporally situated (i.e. the existential restriction on the property \texttt{:atTime}), allowing for time-based queries and sequencing.

At the action level, the class \texttt{:Action} represents a concrete activity carried out by an agent within the environment and is formalised as a subclass of \texttt{dul:Action}. Unlike \texttt{:Task}, which models the abstract or planned specification of an activity, \texttt{:Action} captures the actual execution or occurrence of that activity. Hence, an \texttt{:Action} is related to: (i) the task it realises via the property \texttt{:isExecutionOf}, (ii) the agent who performs it via the property \texttt{:isPerformedBy}, and (iii) the resulting world state it produces via the property \texttt{:bringsAbout}. 

By modelling \texttt{:PlanExecution} and \texttt{:Action} explicitly, the ontology supports reasoning over whether, how, and to what extent plans have been carried out, enabling comparisons between intended and actual behaviour. It also provides a foundation for temporal monitoring, traceability, and post-hoc analysis of agent decisions, which is crucial for explainability and accountability in both human and artificial agents.

\paragraph{\bf Temporal reasoning.}
A fundamental principle underpinning our modelling is the concept of time. Each class in the ontology can be associated with a Time entity that indicates the start and end of the validity of an instance of any given class. 

\begin{align}
& \textit{TemporalEntity} \sqsubseteq \textit{dul:Region} \label{dl:time_s}\\
& \textit{TimeInstant} \sqsubseteq \textit{TemporalEntity} \\
& \textit{TimeInterval} \sqsubseteq \textit{TemporalEntity} \\
& \textit{TimeInterval} \equiv \textit{dul:TimeInterval} \\
& \textit{TimeInterval} \sqsubseteq ({=}1 \textit{ hasStartTime.TimeInstant}) \\
& \textit{TimeInterval} \sqsubseteq ({\leq}1 \textit{ hasEndTime.TimeInstant}) \\
& \textit{MentalEntity} \sqsubseteq \exists \textit{atTime.TemporalEntity} \\
& \textit{MentalEntity} \sqsubseteq \exists \textit{hasValidity.TemporalEntity}
\label{dl:time_e} 
\end{align}

This facilitates the monitoring of temporal changes in the elements of the system, enabling the identification of alterations undergone by different agents. The ontology defines the class \texttt{:TemporalEntity} as a subclass of \texttt{dul:Region}. Hence, temporal constructs in the ontology are treated as abstract value spaces rather than as concrete events or processes. This modelling choice enables the ontology to represent temporal aspects as values of temporal qualities and facilitates integration with other DUL-compliant ontologies that rely on the region-based interpretation of time. The ontology distinguishes two primary types of temporal constructs as subclasses of \texttt{:TemporalEntity}: (i) \texttt{:TimeInstant} and (ii) \texttt{:TimeInterval}. A \texttt{:TimeInstant} represents a zero-duration temporal point, typically used to indicate a specific moment in time (e.g. a timestamp or a boundary between intervals). It serves as a temporal marker that can be used to define the start or end of a temporal interval. Conversely, a \texttt{:TimeInterval} denotes a temporally extended region, bounded by a start and optionally an end point in time. Formally, a \texttt{:TimeInterval} is constrained to have exactly one starting \texttt{:TimeInstant} and at most one ending \texttt{:TimeInstant}, reflecting both open and closed intervals. This distinction enables the ontology to capture both instantaneous and durative aspects of temporal reasoning, and aligns with well-established patterns in temporal ontologies. Every mental entity is temporally located or constrained either by being associated with a specific time (i.e. through the object property \texttt{:atTime}) or by having a temporal validity condition (i.e. through the object property \texttt{:hasValidity}). The use of both characterisations supports alternative but complementary views. In fact, on the one hand, \texttt{:atTime} captures temporal anchoring, useful for representing when a mental state or process occurs, on the other \texttt{:hasValidity} captures temporal extent or applicability, such as the duration over which a belief is held or an intention remains active. Modelling the class \texttt{MentalEntity} through existential restrictions on these properties ensures that all mental constructs in the ontology are temporally contextualised, enabling time-aware reasoning over agents' cognitive dynamics and supporting alignment with temporal patterns in upper ontologies.

\end{document}